# An Online Character Recognition System to Convert Grantha Script to Malayalam

Sreeraj.M
Department of Computer Science
Cochin University of Science and Technology
Cochin-22, India

Sumam Mary Idicula
Department of Computer Science
Cochin University of Science and Technology
Cochin-22, India

*Abstract*— **This paper presents a novel approach to recognize Grantha, an ancient script in South India and converting it to Malayalam, a prevalent language in South India using online character recognition mechanism. The motivation behind this work owes its credit to (i) developing a mechanism to recognize Grantha script in this modern world and (ii) affirming the strong connection among Grantha and Malayalam. A framework for the recognition of Grantha script using online character recognition is designed and implemented. The features extracted from the Grantha script comprises mainly of time-domain features based on writing direction and curvature. The recognized characters are mapped to corresponding Malayalam characters. The framework was tested on a bed of medium length manuscripts containing 9-12 sample lines and printed pages of a book titled Soundarya Lahari writtenin Grantha by Sri Adi Shankara to recognize the words and sentences. The manuscript recognition rates with the system are for Grantha as 92.11%, Old Malayalam 90.82% and for new Malayalam script 89.56%. The recognition rates of pages of the printed book are for Grantha as 96.16%, Old Malayalam script 95.22% and new Malayalam script as 92.32% respectively. These results show the efficiency of the developed system.**

## I. INTRODUCTION

Analysis of handwritten data using computational techniques has been accelerated with the growth of computer science developing human-computer interaction. To obtain handwritten data in digital format, the writing can be scanned or the writing itself can be done with the aid of special pen interfaces. The two techniques are commonly known as off-line and on-line handwriting respectively. Offline handwriting recognition focuses recognition of characters and words that had been recorded earlier in the form of scanned image of the document. In contrast, online handwriting recognition focuses on tasks when recognition can be performed at the time of writing through the successive points of strokes of the writer in a fraction of time.

This paper is an attempt to perform online character recognition for Grantha script and is the first of its kind to the best of our knowledge. Grantha script is an ancient language evolved from Brahmic script. The Dravidian-South Indian-languages have succeeded Grantha and Brahmi. Many of our ancient literature are in Grantha. Extract the knowledge from this extinct language is difficult. Grantha has a strong linkage with Malayalam characters, so our work uses this similarity to convert the Grantha script into Malayalam. This boosts our system into extra mile.

The recognition of handwritten characters in Grantha script is quite difficult due to the numerals, vowels, consonants, vowel modifiers and conjunct characters. The structure of the scripts and the variety of shapes and writing style of individuals at different times and among different individual poses challenges that are different from the other scripts and hence require customized techniques for feature representation and recognition. Selection of a feature extraction method is the single most important factor in achieving high recognition performance [1]. In this paper a framework for recognizing Grantha script and mapping to its corresponding Malayalam character is described.

The paper is organized as follows. Section 2 gives the related work in the field of Indic scripts. Section 3 portrays an overview of the Grantha Script. Section 4 details the features of Malayalam language whereas Section 5 points to the snaps of linkage between Grantha Script and Malayalam. Section 6 depicts the framework for recognizing Grantha script by converting to Malayalam using online character recognition mechanism. Section 7 describes the feature extraction techniques adopted in this framework. Section 8 explains implementation details and Section 9 analyses and discusses the experiments and their results. The paper is concluded in Section 10.

## II. RELATED WORK

Many works have been done in linguistics and literature focusing on the recognition of simple characters–independent vowels and isolated consonants. There are three well studied strategies for recognition of isolated (complex) characters for scripts like Devanagari, Tamil, Telugu, Bangla and Malayalam. (i) Characters can be viewed as composition of strokes.[2],[3],[4],[5],[6] (ii) Characters may be viewed as compositions of C, C', V and M graphemes [7] (iii) character can be viewed as indivisible units. [8].

## III. OVERVIEW OF GRANTHA SCRIPT

The Grantha script is evolved from ancient Brahmic script and it has parenthood of most of the Dravidian-south Indian-languages. In Sanskrit, 'Grantha' stands for 'manuscript'. In 'Grantha', each letters represents a consonant with an inherent vowel (a). Other vowels were indicated using a diacritics or separate letters. Letters are grouped according to the way they are pronounced. There are 14 vowels. Of these 7 are the basic symbols. Long vowels and diphthongs are derived from these. Also there exists 13 vowel modifiers, and there are no full vocalic short l and full vocalic long l modifier. 'Grantha' admits 34 basic consonant characters. As with all Brahmi derived scripts, the consonant admits the implicit vowel 'schwa'. Pure consonant value is obtained by use of the virāma. 'Grantha' has two diacritic markers: the anuswāra (ⓞ) and the visarga (ⓢ). The anuswāra is a latter addition and in Archaic as well as Transition 'Grantha' the letter ma is used to represent the nasal value. A special feature of 'Grantha' is the use of subsidiary symbols for consonants. These are three in number: the use of a subsidiary ya and two allographs for ra depending on whether ra precedes the consonant or follows it [9]. The Fig.1 gives the symbols used in 'Grantha' script.

The consonant ௱ is represented in two ways. When following a consonant it is written as ⌐ under the consonant; but when it precedes a consonant it takes the ⌐ form written after the consonant or conjunct.

Complex consonantal clusters in 'Grantha' script use the Samyuktaksaras (conjunct letters) very widely. The Samyuktaksaras of 'Grantha' is formed in the following three ways [10]. They are stacking, combining and using special signs as shown in the following Table I. Combined with vowel signs, these Samyuktaksaras are considered as a single unit and placed with the Vowel signs.

TABLE I. FORMATION OF SAMYUKTAKSARAS (CONJUNCT LETTERS) IN GRANTHA SCRIPTS

| Stacking | | ஷ்+ப்+ய்+வ→ ஷ்+ப்ய → ஷ்ப்ய |
|---|---|---|
| Combining | | க்+ஷ → க்ஷ |
| Special signs | -r- Conjunct | நி்+ய்+ா→ ந்ய்+ா→ந்யா, ரி்+ா+உ→ ரி்+ய்க→ ய்காு |
| | -y- Conjunct | தி்+ய→தய், ய்+கூ→ யூ, நி்+தி்+ய→ கி்+ய→ய |

IV. MALAYALAM LANGUAGE

Malayalam is a Dravidian language consisting of syllabic alphabets in which all consonants have an inherent vowel. Diacritics are used to change the inherent vowel and they can be placed above, below, before or after the consonant. Vowels are written as independent letters when they appear at the beginning of a syllable. Special conjunct symbols are used to combine certain consonants. There are about 128 characters in the Malayalam alphabet which includes Vowels (15), consonants (36), chillu (5), anuswaram, visargam, chandrakkala-(total-3), consonant signs (3), vowel signs (9), and conjunct consonants (57). Out of all these characters mentioned, only 64 of them are considered to be the basic ones as shown in Fig. 2.

The properties of Malayalam characters are the following

Since Malayalam script is an alphasyllabary of the Brahmic family they are written from left to right.

- Almost all the characters are cursive by themselves. They consist of loops and curves. The loops are written frequently in the clockwise order

- Several characters are different only by the presence of curves and loops.

- Unlike English, Malayalam scripts are not case sensitive and there is no cursive form of writing.

Malayalam is a language which is enriched with vowels, consonants and has the maximum number of sounds that are not available in many other languages as shown in Fig. 3.

**Vowels**

| அ | ஆ | இ | ஈ | உ | ஊ |
|---|---|---|---|---|---|
| a | ā | i | ī | u | ū |

| ஃ | ஃ | எ | ஏ3 | | |
|---|---|---|---|---|---|
| ṛ | ṝ | ḷ | ḹ | | |

| ஹ | ஹை | ஒ | ஔ | | |
|---|---|---|---|---|---|
| e | ai | o | au | | |

| o | ஃ |
|---|---|
| ṁ | ḥ |

**Consonants**

| க | ஖ | ஗ | ஘ | ங |
|---|---|---|---|---|
| k | kh | g | gh | ṅ |

| ச | ச | ஜ | ஜ | ஞ |
|---|---|---|---|---|
| c | ch | j | jh | ñ |

| ட | ட | ட | ட | ண |
|---|---|---|---|---|
| ṭ | ṭh | ḍ | ḍh | ṇ |

| த | த | த | த | ந |
|---|---|---|---|---|
| t | th | d | dh | n |

| ப | ப | ப | ப | ம |
|---|---|---|---|---|
| p | ph | b | bh | m |

| ய | ர | ல | வ | ள |
|---|---|---|---|---|
| y | r | l | v | ḻ |

| ஶ | ஷ | ஸ | ஹ |
|---|---|---|---|
| ś | ṣ | s | h |

Figure 1. Grantha characters

Figure 2. 64 basic characters of Malayalam

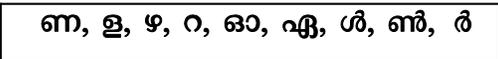

Figure 3. Rare Sounds of scripts available only in Malayalam language

## V. GRANTHA SCRIPT AND MALAYALAM – SNAPS OF LINKAGE

The foundation of Malayalam script owes itself to Grantha script. They have much similarity with Grantha scripts. When Grantha scripts were used to write Sanskrit sounds/phonemes, it was called Kolezhuthu (rod script).

### A. Challenges between Grantha and old scripts of Malayalam

1. Complex stacking conjuncts up to Triple conjuncts are present in Grantha and it is up to two conjuncts in the old script of Malayalam.

2. Complex Combining Conjuncts with 4 (or more) consonantal clusters are present in Grantha but in old Malayalam script it is only up to 2 consonantal clusters.

### B. Challenges between Grantha and new scripts of Malayalam

1. The special vowelless forms of the Grantha consonants ഇ & ൪ , ഇ & ൪ rarely in printings. This sort of consonants including this can be seen in manuscripts.

2. Number of vowels is decreased by the absence of characters corresponding to ഋ(ൃ),ൠ(ൄ),ഌ(ൢ)

3. Complex stacking conjuncts are present in Grantha and it is absent in the new script of Malayalam.

4. Complex Combining Conjuncts with 4 (or more) consonantal clusters are present in Grantha but in new Malayalam script it is only up to 2 consonantal clusters.

## VI. SYSTEM FRAMEWORK

The framework designed comprises of five modules. The first module preprocesses the datasets, to necessitate the way for feature extraction. The second module is the feature extraction module. The features extracted here are time domain features based on writing direction and curvature, which is discussed in detail in the following section. The next two modules are part of the classification process namely trainer and recognizer. The knowledge feature vector of the model data from the trainer is fed as one of the inputs to the recognizer in the testing phase. The recognizer is aided with the Grantha conjugator, which could extract the rules from conjunct characters. The fifth module is the converter, where the conversion of Grantha script to old and new Malayalam characters is done. The converter is supported by intellisense feature, which is incorporated to avoid the problem corresponding to the absence of certain characters in new script of Malayalam by providing the equivalent word by searching from a dictionary. Also Malayalam conjugator aided the converter to form rules from conjunct characters to be converted to new scripts of Malayalam. The entire framework is presented in the following Fig.4.

## VII. FEATURE EXTRACTION

This is the module where the features of handwritten characters are analyzed for training and recognition which are explained below. In proposed approach, each point on the strokes with values of selected features (time-domain features [11], [12] writing direction and curvature) are described in the consecutive sub sections.

### A. Normalized x-y coordinates

The x and y coordinates from the normalized sample constitute the first 2 features.

### B. Pen-up/pen-down

In this system the entire data is stored in UNIPEN format where the information of the stroke segments are exploited using the pen-up/pen-down feature. Pen-up/pen-down feature is dependent on the position sensing device. The pen-down gives the information about the sequence of coordinates when the pen touches the pad surface. The pen-up gives the information about the sequence of coordinates when the pen not touching the pad surface.

### C. Aspect ratio

Aspect at a point characterizes the ratio of the height to the width of the bounding box containing points in the neighborhood. It is given by

$$A(t) = \frac{2 \times \Delta y(t)}{\Delta x(t) + \Delta y(t)} - 1 \qquad (1)$$

Where $\Delta x(t)$ and $\Delta y(t)$ are the width and the height of the bounding box containing the points in the neighborhood of the point under consideration. In this system, we have used neighborhood of length 2 i.e. two points to the left and two points to the right of the point along with the point itself.

### D. Curvature

The curvature at a Point (x (n), y (n)) is represented by cosφ (n) a and sinφ (n). It can be computed using the following formulae [11].

$$Sin\phi(n) = Cos\theta(n-1) \times Sin\theta(n+1) - Sin\theta(n-1) \times Cos\theta(n+1) \qquad (2)$$

$$Cos\phi(n) = Cos\theta(n-1) \times Cos\theta(n+1) + Sin\theta(n-1) \times Sin\theta(n+1) \qquad (3)$$

### E. Writing direction

The local writing direction at a point (x (n), y (n)) is described using the cosine and sine [12].

$$\sin\theta(n) = \frac{Y_n - Y_{n-1}}{\sqrt{(X_n - X_{n-1})^2 + (Y_n - Y_{n-1})^2}} \qquad (4)$$

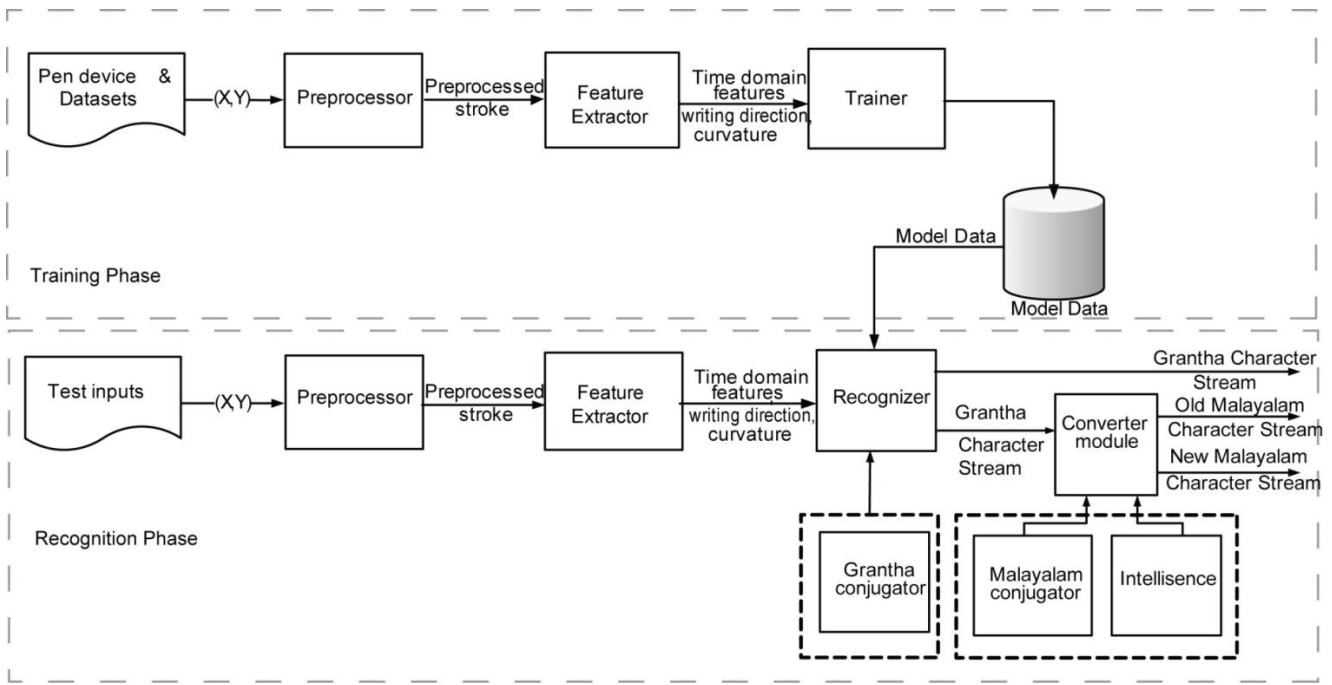

$$\cos\theta(n) = \frac{X_n - X_{n-1}}{\sqrt{(X_n - X_{n-1})^2 + (Y_n - Y_{n-1})^2}} \quad (5)$$

Figure 4. System Architecture

The above elements will be the feature vector for training and recognition modules. The classifier used here is k-NN. Dynamic Time Warping DTW distance is used as the distance metric in the k-Nearest Neighbor classifier. Dynamic Time Warping is a similarity measure that is used to compare patterns, in which other similarity measures are practically unusable. If there are two sequences of length n and m to be aligned using DTW, first a nXm matrix is constructed where each element corresponds to the Euclidean distance between two corresponding points in the sequence. A warping path W is a contiguous set of matrix elements that denotes a mapping between the two sequences. The W is subject to several constraints like boundary conditions, continuity, monotonicity and windowing. A point to point correspondence between the sequences which satisfies constraints as well as of minimum cost is identified by the following eqn.6.

$$DTW(Q,C) = \min\left\{\sqrt{\sum_{k=1}^{K} w_k}\bigg/K\right\} \quad (6)$$

Q, C are sequences of length n and m respectively. Wk element is in the warping path matrix. The DTW algorithm finds the point-to-point correspondence between the curves which satisfies the above constraints and yields the minimum sum of the costs associated with the matching of the data points. There are exponentially many warping paths that satisfy the above conditions. The warping path can be found efficiently using dynamic programming to evaluate a recurrence relation which defines the cumulative distance (i, j) as the distance d(i, j) found in the current cell and the minimum of the cumulative distances of the adjacent elements. [13].

## VIII. IMPLEMENTATION

The online character recognition is achieved through two steps i) Training and ii) Recognition. Training of the samples is done setting the prototype selection as hierarchical clustering. Recognition function predicts the class label of the input sample by finding the distance of the sample to the classes according to the K-Nearest Neighbor rule. DTW distance is used as the distance metric in the k-Nearest Neighbor classifier. The top N nearest classes along with confidence measures are returned as the identified letters.

Conversion from Grantha to Malayalam words are done in two ways. They are converted to old and new scripts of Malayalam. Due to the challenges posed (section V.(A)) the conversion between Grantha and New scripts of Malayalam is a very tedious task when compared with the old Malayalam script. So we adopted the Malayalam conjugator. Following Algorithm 1 explain the conversion of Grantha word to Malayalam old and new script.

**Algorithm 1.** Conversion of Grantha word to Malayalam old and new script

Input: **Grantha Word $W_g$**

Output: **Malayalam Word $W_{mo}$ and $W_{mn}$ in old script and new script respectively**

for **each character $w_g(i)$ in $W_g$**

    if ($w_g(i) == $ ൂ )

    then

        *temp_store = $w_g(i-1)$;*

        *$w_g(i-1) = w_g(i)$;*

```
            w_g(i) = temp_store;
        end
    end
    for each character w_g(i) in W_g
        Set char_type;
        if char_type == vowel or char_type == consonant
            Set w_mo(i) and w_mn(i) directly
        else if char_type == complex stacking form of conjunct letters
            Find R from conjugator// example of R=
            ஸ்ட→ஃ+ஸ்→ஃ+ட்+ஸ்
            Set w_mo(i) and w_mn(i)
        else if char_type == combined form of conjunct letters
            Find R' from conjugator //example of R'=
            க்தூ→வ+த்→வ+த்+ூ
            Set w_mo(i) and w_mn(i)
        End
    W_mo = Concatenate (w_mo(i));
    W_mn = Concatenate (w_mn(i));
    End
```

## IX. EXPERIMENTAL RESULTS AND DISCUSSION

Medium length manuscripts containing 9-12 sample lines of which 34-42 characters are in each line varied which are collected to perform experiments. So, in effect, in each manuscript the number of characters are varied from 306-504. Experiment is conducted to evaluate the character recognition system. The collection of basic characters which can be used to make all the characters of Grantha script is formed by a rule based system. It is understood that some characters are frequently misclassified. So experiments are conducted on those characters to find their similarity measure and Confusion matrix of such characters are shown in Fig 5. The correct word limiter space in manuscript was not able to be recognized and it is resolved by searching the possible word from the dictionary while making the conversion between Grantha and Malayalam. Test is conducted to recognize 26712 symbols and 3180 words from the manuscript.

Test is also conducted on the printed pages of the Grantha copy of the book titled 'Soundarya Lahari' to recognize the words and sentences. Each page consists of 32-35 lines and words vary from 160-190. The test is conducted only to recognize 455 lines due to the availability of limited number of pages. Table 2 summarizes the result of recognition rate of words. The recognition rate for the different symbols in Grantha Script was tried with classifiers like k-NN and SVM. Specific experiments were done with and without DTW algorithm using k-NN classifier. A bar graph is plotted in Fig 6 with different classifiers and recognition rates.

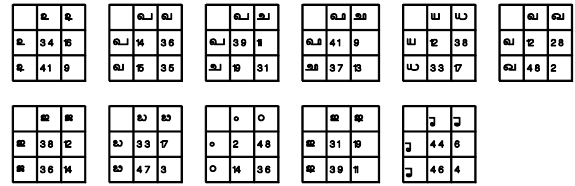

Figure 5. Confusion Matrix of Frequently misclassified characters

TABLE II. RECOGNITION RATE OF GRANTHA WORDS

|            | Grantha | old Malayalam | new Malayalam |
|------------|---------|---------------|---------------|
| manuscript | 92.11%  | 90.82%        | 89.56%        |
| book       | 96.16%  | 95.22%        | 92.32%        |

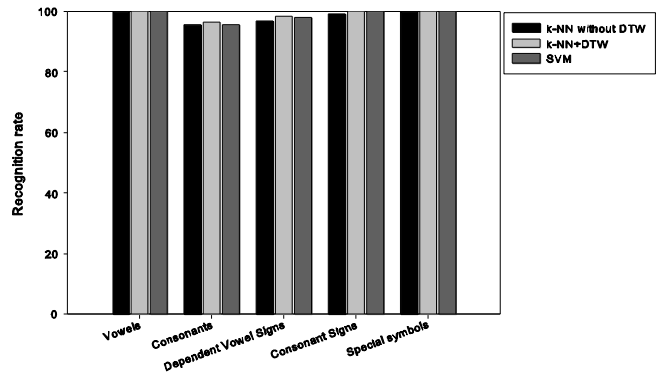

Figure 6. Category-wise comparison of recognition rates using different classifiers

## X. CONCLUSION

This paper presented a framework for recognizing Grantha script, by converting to Malayalam using online character recognition mechanism. The framework was designed with different modules to perform the same. The feature extraction module was designed explicitly by taking into consideration of the different features of Grantha script, which is discussed in detail in this paper. The framework was implemented and tested with manuscripts as well as a book. Experiments were conducted to analyze the different phases of the framework. Different distance measures for evaluating the similarity were tried with, of which DTW method showed considerable recognition rates. It has been observed that there are frequently misclassified characters. They have been studied in detail with the aid of confusion matrix Also the recognition rate was evaluated against two classifiers, namely, k-NN and SVM classifier. Polynomial, sigmoid and RBF kernel functions were used for testing with SVM classifier.

REFERENCES

[1] O. D. Trier, A. K.Jain and T. Taxt,. Feature Extraction Methods For character Recognition : A survey. J Pattern Recognition 1999, 29, 4, 641 – 662.

[2] H. Swethalakshmi, Online Handwritten Character recognition for Devanagari and Tamil Scripts using Support Vector Machines, Master's thesis, Indian Institute of Technology, Madras, India: 2007.

[3] A. Jayaraman, C. C. Sekhar andV. S. Chakravarthy,. Modular Approach to Recognition of Strokes in Telugu Script. In proceedings of 9th Int.

Document Analysis and Recognition Conference (ICDAR 2007), Curitiba, Brazil: 2007.

[4] K. H. Aparna, V. Subramanian, M. Kasirajan, G.V. Prakash, V. S. Chakravarthy and Madhvanath, S. Online Handwriting Recognition for Tamil. In proceedings of 9th Int. Frontiers in Handwriting Recognition Workshop (IWFHR 2004), Tokyo, Japan: 2004.

[5] M. Sreeraj and Sumam mary Idicula. On-Line Handwritten Character Recognition using Kohonen Networks. In Proceeding IEEE 2009 World Congress on Nature & Biologically Inspired Computing (NABIC'09), Coimbatore,India: 2009;1425-1430..

[6] M. Sreeraj and Sumam mary Idicula. k-NN Based On-Line Handwritten Character Recognition System. In proceedings of First International Conference on Integrated Intelligent Computing (iciic 2010), IEEE, Bangalore: 2010;171-176.

[7] V.J. Babu, L. Prashanth, R.R. Sharma, G.V.P. Rao, and Bharath, A. HMM-Based Online Handwriting Recognition System for Telugu Symbols. In Proceeding of the.9th International Document Analysis and Recognition Conference  (ICDAR 2007), Curitiba, Brazil: 2007.

[8] N. Joshi, G. Sita, A.G. Ramakrishnan, V. Deepu and , S. Madhvanath. Machine Recognition of Online Handwritten Devanagari Characters. In proceedings of 8th International Document Analysis and Recognition Conference  (ICDAR 2005), Seoul, Korea: 2005.

[9] http://tdil.mit.gov.in/pdf/grantha/pdf_unicode_proposal-grantha_.pdf

[10] http://www.virtualvinodh.com/grantha-lipitva

[11] M. Pastor,  A.Toselli and  E. Vidal. Writing Speed Normalization for On-Line Handwritten Text Recognition. In proceedings of the International Conference on Document Analysis and Recognition (ICDAR). 2005.

[12] I. Guyon, P. Albrecht, Y. Le Cun, J. Denker and W. Hubbard, Design of a Neural Network Character Recognizer for a Touch Terminal. J Pattern Recognition 1991,  24(2):105–119.

[13] N. Joshi,  G. Sita,  A.G. Ramakrishnan and , S. Madhvanath. Tamil Handwriting Recognition Using Subspace and DTW Based Classifiers. In proceedings of 11th International Neural Information Processing Conference (ICONIP 2004), Calcutta, India: 2004; 806-813.